\documentclass[letterpaper, 10 pt, conference]{ieeeconf}  

\IEEEoverridecommandlockouts                              

\usepackage{graphics} 
\usepackage{epsfig} 
\usepackage{mathptmx} 
\usepackage{times} 
\usepackage{amsmath} 
\usepackage{amssymb}  
\usepackage{subfigure}
\usepackage{algorithmicx}
\usepackage{algpseudocode,algorithm}
\usepackage{color}
\usepackage{comment}
\usepackage{url} 
\definecolor{hr_color}{rgb}{1.0, 0.0, 0.0}

\title{\LARGE \bf
Magnetic Field Aided Vehicle Localization with Acceleration Correction
}

\author{Mrunmayee Deshpande$^{1}$, Manoranjan Majji$^{2}$ and J. Humberto Ramos$^{3}$ 
\thanks{$^{1}$Mrunmayee Deshpande is a PhD Student at the Aerospace Engineering Department, Texas A\&M University, College Station, TX, USA.
        {\tt\small mdesh@tamu.edu}}%
\thanks{$^{2}$Manoranjan Majji is an Associate Professor at the Aerospace Engineering Department, Texas A\&M University, College Station, TX, USA.}%
\thanks{$^{2}$J. Humberto Ramos is a Research Assistant Scientist at the Department of Mechanical and Aerospace Engineering at the University of Florida, Fl, USA.}%
}

\begin{document}

\maketitle
\thispagestyle{empty}
\pagestyle{empty}

\begin{abstract}
This paper presents a novel approach for vehicle localization by leveraging the ambient magnetic field within a given environment. Our approach involves introducing a global mathematical function for magnetic field mapping, combined with Euclidean distance-based matching technique for accurately estimating vehicle position in suburban settings. The mathematical function based map structure ensures efficiency and scalability of the magnetic field map, while the batch processing based localization provides continuity in pose estimation. Additionally, we establish a bias estimation pipeline for an onboard accelerometer by utilizing the updated poses obtained through magnetic field matching. Our work aims to showcase the potential utility of magnetic fields as supplementary aids to existing localization methods, particularly beneficial in scenarios where Global Positioning System (GPS) signal is restricted or where cost-effective navigation systems are required. 
\end{abstract}

\section{INTRODUCTION}

Autonomous navigation has been an extensively studied field over the past two decades. Most autonomous systems make use of GPS, Light Detection and Ranging (LiDAR), and Camera as sensory inputs to estimate their position in a particular environment, as elaborated in \cite{SurveyLocalization}. While these sensors and their combinations yield promising outcomes \cite{lidar-gnss}, \cite{lidar-cam}, \cite{Lidar}, the imperative to explore alternative or supplementary options remains paramount. Environments with restricted GPS signal or feature deficit surroundings present significant challenges for traditional localization systems. Additionally, the need for cost-effective systems encourages to re-examine usability of readily available on-board sensors like magnetometers. This impetus drives our investigation into utilizing the ambient magnetic field to augment vehicle localization in suburban settings.

The magnetic field based localization has been studied extensively for indoor applications. The ambient magnetic field offers a robust feature-based map for indoor localization, addressing the limitations of conventional positioning systems such as GPS within enclosed environments \cite{ramos2022information,ramos2024ownship}. In \cite{survey}, a comprehensive survey shows the diverse approaches explored in the context of mapping and leveraging the ambient magnetic field for indoor localization. Additionally, in \cite{Indoor1}, a particle filtering technique tailored for indoor localization using a magnetic one-dimensional map is shown to accurately estimate a robot's position along indoor corridors. Along with these studies, there are some interesting studies done for the outdoor environment as well. For instance, \cite{Outdoor1} employs a particle filter technique to achieve localization utilizing a pre-existing magnetic field map. This method estimates longitudinal travel distance and utilizes azimuth angle errors to gauge lateral deviations. Similarly, \cite{Outdoor2} explores outdoor localization leveraging ambient magnetic fields with wheel encoders to determine azimuth angles and traveled distances. Magnetic field-based localization presents a viable solution for navigating GPS-denied areas such as tunnels, underground parking facilities, and multi-level roads, as illustrated in \cite{MagLand}. The promising results shown in these studies offer a possible solution to enhance outdoor localization capabilities, particularly in challenging environments where traditional navigation systems may be unreliable.

Our research introduces a methodology for combining Inertial Measurement Unit (IMU) based pose predictions with efficient mapping and localization using magnetic field data. The method utilizes batches of magnetic field data for a continuous pose estimation $(x,y)$ instead of relying on magnetic field spikes through the environment. The method utilizes the positions updated through magnetic field matching to infer accelerometer bias and scale factor. We validate the effectiveness of our approach by testing through on-road applications, employing an IMU and a magnetometer positioned within a vehicle while driving in suburban environments. 
\section{System Overview}
\begin{figure}
\centering
\includegraphics[scale = 0.38]{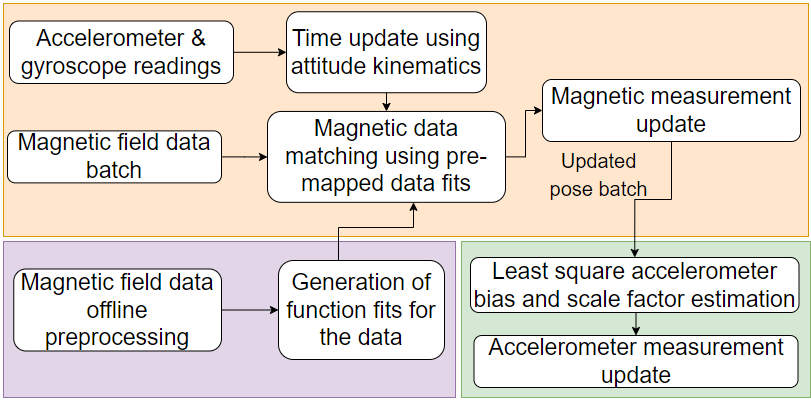}
\caption{System overview representing the magnetic field aided localization along with magnetic field data mapping and accelerometer bias and scale factor estimation }
\label{fig:overview}
\end{figure}
The system architecture proposed for magnetic field aided localization is depicted in Fig. \ref{fig:overview}. The system is mainly comprised of four main sub-systems. The pre-processing of the data, time update or state prediction from the IMU measurements, magnetic field data matching based pose estimation and least square based accelerometer bias and scale factor estimation. The ambient magnetic field data is processed offline to create a map structure for localization. Gyroscope and accelerometer data is processed in real-time for predicting the vehicle poses, using attitude kinematics. The magnetic field data obtained in real-time is processed in batches and compared against the pre-built map for estimating the vehicle pose. Along with the magnetic field data based update, accelerometer bias and scale factor are obtained. The algorithm utilizes Extended Kalman Filter (EKF) to estimate the vehicle's final pose. Algorithm [\ref{alg:EKF}] details the implementation of EKF designed to integrate the estimations of the sub-systems.


\section{Magnetic Field Map} \label{Data Representation}
\subsection{Linear Coordinate System}
In addressing the challenge of localization, maintaining an efficient map structure is imperative for achieving fast and reliable real-time lookup capabilities. We propose utilizing three unidimensional map structures based on the distance traveled by the vehicle. This approach facilitates the representation of scalar magnetic field readings, and ground truth $x$ positions and $y$ positions of the vehicle with respect to cumulative distance $S$ traveled by the vehicle as shown in Fig. \ref{fig:MagMatch}. The cumulative distance $S$ is the distance covered by the vehicle at each time step, starting from the initial time step. Considering arbitrary positions of a moving vehicle where magnetic field readings are obtained: \(x_1,y_1;x_2,y_2;...;x_n,y_n\), the corresponding $S$ values at each time step are determined as follows:
\begin{equation}
\begin{split}
s_1 & = 0 \\
s_n & = s_{n-1} + \sqrt{(x_n-x_{n-1})^2 + (y_n-y_{n-1})^2}
\end{split}
\end{equation} 

\subsection{Data Function Fits}
\begin{figure}
\centering
\includegraphics[scale=0.45]{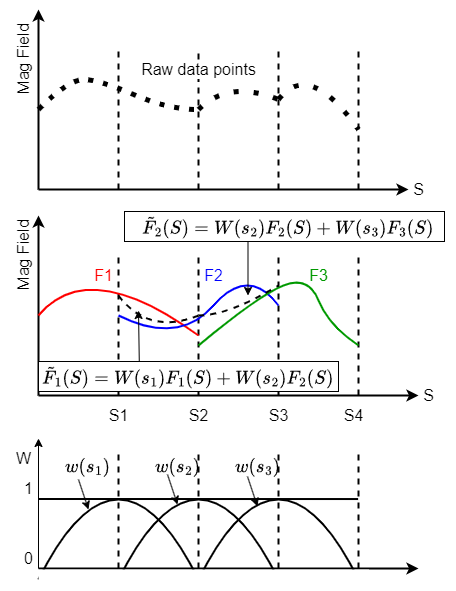} 
\caption{Function fit using weighting technique: Local function fits are merged together by utilizing the weighting scheme.}
\label{fig:glomap}
\end{figure}
In order to make an efficient map structure, the data obtained in the previous step is partitioned into equal-length segments of traveled distances $h$. Within each consecutive pair of segments, a local third-degree Legendre polynomial function is fitted. The synthesis of global fit functions is achieved through a weighting algorithm known as $Global \ Mapping \ (GLOMAP)$, detailed in \cite{glomap}. The merging of consecutive local fits is facilitated by a weighting function, which is articulated as:
\begin{equation}
  W(s)=\begin{cases}
    1-s^2(3+2s), & \text{$-1\leq s\leq 0$}\\
    1-s^2(3-2s), & \text{$0\leq s\leq 1$}
  \end{cases}
\end{equation}
As illustrated in Fig. \ref{fig:glomap}, the $F_1, F_2, F_3$ represent local fits while $\tilde{F}_1, \tilde{F}_2$ represent the global fits. The weighting function, $w$, ranges from 0 to 1 such that the shifted weight functions always equate to 1. This function-fit approach allows us to efficiently store the function coefficients rather than the entire data.  
\section{System Kinematics and Filtering} \label{Attitude Kinematics}
The extent of the current study is to estimate a 2D pose of the vehicle along with accelerometer biases. However, in order to lay the foundation to include a 3D attitude estimation for future work, the attitude kinematics of the system is based on the following state vector: 
\begin{equation} \label{eq:stateVec}
    X = [q, p, v, \beta_g,\beta_a, k_g, k_a]^T
\end{equation}
Here, $q$ is a quaternion representing the orientation of the vehicle, $p$ is the position vector of the vehicle in a local reference frame, $v$ is the velocity vector in a local reference frame, $\beta_g$ and $K_g$ are the bias and scale factor of the gyroscope and $\beta_a$ and $K_a$ are the bias and scale factor of the accelerometer. Note that, $k_g$ and $k_a$ are the elements of the diagonal matrices $K_g$ and $K_a$. The measurement models of the gyroscope and accelerometer are considered as follows:
\begin{align}
\omega&=(I_{3 \times 3}-K_g)(\Tilde{\omega} - \beta_g) - \eta_{gv} & \dot{\beta_g}&=\eta_{gu}  \nonumber \\ 
a&=K_a\Tilde{a} - \beta_a - \eta_{av} & \dot{\beta_a}&=\eta_{au}
\end{align}
where, $\eta_{gv}$, $\eta_{gu}$, $\eta_{av}$ and $\eta_{au}$ are zero-mean Gaussian random variables. The corresponding estimation models are:
\begin{align} \label{eq:estimationModel}
\hat{\omega}&=(I_{3\times3}-K_g)(\Tilde{\omega} - \hat{\beta}_g) & \dot{\hat{\beta}}_g&=0 & \dot{\hat{k}}_g &= 0\nonumber\\ 
\hat{a}&=K_a\Tilde{a} - \hat{\beta}_a & \dot{\hat{\beta}}_a&=0 & \dot{\hat{k}}_a &= 0 
\end{align}
Note that the accelerometer data is to be transformed into the local frame of reference utilizing the orientation data acquired from the car's gyroscope measurements. Leveraging the IMU data, the propagation of the states from time step $k$ to $k+1$, where $\Delta t$ is the time between the two time steps, is implemented as follows:
\begin{align} 
\label{timeUpdate}
    \hat{q}_{k+1} &= \Phi(\hat{\omega}_k)\hat{q}_k \\ \nonumber
    \hat{p}_{k+1} &= \frac{1}{2} C(\hat{q}_k)^T \hat{a}_k \Delta t^2 + \hat{v}_k \Delta t + \hat{p}_k \\ \nonumber
    \hat{v}_{k+1} &= C(\hat{q}_k) \hat{a}_k \Delta t + \hat{v}_k \\ \nonumber
    \hat{\beta}_{g_{k+1}} &= \hat{\beta}_{g_k} \\ \nonumber
    \hat{\beta}_{a_{k+1}} &= \hat{\beta}_{a_k} \\ \nonumber
    \hat{k}_{g_{k+1}} &= \hat{k}_{g_k} \\ \nonumber
    \hat{k}_{a_{k+1}} &= \hat{k}_{a_k}
\end{align}
Here, $C(\hat{q_k})$ is direction cosine matrix and $\Phi(\hat{\omega}_k)$ is given as \cite{crassidis2004optimal},
\begin{align}
    \Phi(\hat{\omega}_k) &= \begin{bmatrix}
        cos(\frac{1}{2} ||\hat{\omega}_k|| \Delta t)I_3 - [\hat{\psi}_k \times] & \hat{\psi}_k \\
        -\hat{\psi}^T_k & cos(\frac{1}{2} ||\hat{\omega}_k|| \Delta t)
    \end{bmatrix} \nonumber \\
    \hat{\psi}_k &= \frac{sin(\frac{1}{2} ||\hat{\omega}_k|| \Delta t) \hat{\omega}_k}{||\hat{\omega}_k||}
\end{align}
The state transition matrix $F=\frac{\partial{f(X,\omega,a)}}{\partial{X}}$ for the EKF implementation is obtained as follows:
\begin{equation}
\begin{bmatrix} \begin{smallmatrix}  
-[(\omega \Delta t ) \times] & 0 & 0 & -(I_3 - k_g)\Delta t  & 0 & -(\tilde{\omega}-\beta_g)\Delta t  & 0\\ 
0 & I_3 & I_3 \Delta t  & 0 & -\frac{1}{2}C^T {\Delta t }^2 & 0 & \frac{1}{2}C^T \tilde{a} {\Delta t }^2 \\
0 & 0 & I_3 & 0 & -C^T \Delta t  & 0 & C^T \tilde{a} \Delta t  \\
0 & 0 & 0 & I_3 & 0 & 0 & 0 \\
0 & 0 & 0 & 0 & I_3 & 0 & 0 \\
0 & 0 & 0 & 0 & 0 & I_3 & 0 \\
0 & 0 & 0 & 0 & 0 & 0 & I_3 \\
\end{smallmatrix} \end{bmatrix}  	
\end{equation}
The IMU error vector is given as,
\begin{equation}
    w = [\eta_{gv},\eta_{gu},\eta_{av},\eta_{au}]^T
\end{equation}
The error matrix $G=\frac{\partial{f(X,\omega,a)}}{\partial{w}}$ for the EKF implementation is obtained as follows:
\begin{equation}
\footnotesize
\begin{bmatrix}
-I_3 \Delta t  & 0 & 0 & 0 \\
0 & 0 & -\frac{1}{2}C^T \Delta t^2  & 0 \\
0 & 0 & -C^T \Delta t  & 0 \\
0 & I_3 \Delta t  & 0 & 0 \\
0 & 0 & 0 & I_3 \Delta t  \\
0 & 0 & 0 & 0 \\
0 & 0 & 0 & 0 \\
\end{bmatrix}	
\end{equation}

\section{Magnetic field data matching} 
The magnetic field data is processed in batches for data matching. Let the batch be represented as $M_r$ and global map function be represented as $f_m$, then the Euclidean distance-based data matching is given as:  
\begin{equation} \label{magmatchEQ}
    S_{min} = \underset{S}{\mathrm{argmin}} (\sqrt{[M_r - f_m(S_{i:i+n})]^T[M_r - f_m(S_{i:i+n})]}) 
\end{equation}
To enhance the pose accuracy and to make the process robust, the following steps are taken for performing the magnetic field data matching. 
\label{Mag Matching}
\begin{figure}
\centering
\includegraphics[scale=0.4]{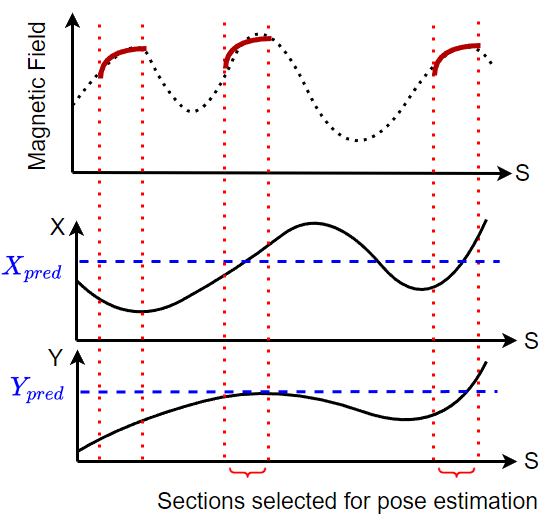}
\caption{Magnetic field data matching technique along with the elimination of false positives. The magnetic field data batches are matched against the function fit map; where the possible matching sections are checked for having intersections with the predicted pose.}
\label{fig:MagMatch}
\end{figure}
\begin{enumerate}
\item As shown in Fig. \ref{fig:MagMatch}, the matching is performed for every incoming batch. The sections of the maps are selected where the matching error is less than a threshold value. As the magnetic field data does not necessarily have unique signatures, multiple selections can be made in this step.
\item The predicted vehicle location $(x_{pred},y_{pred})$ is obtained from the time update step of the EKF, via \eqref{timeUpdate}. Here, $(x_{pred},y_{pred})$ is representing the $\hat{p}_{k+1}$ components. 
\item Using the ($X$ vs $S$) and ($Y$ vs $S$) maps, the sections that intersect with the predicted pose of the vehicle are selected as shown in Fig. \ref{fig:MagMatch}. This allows us to eliminate false positive matches. 
\item The exact $S_{min}$ and the corresponding position $p_{s_{min}} = (x_{s_{min}},y_{s_{min}})$ are obtained by choosing the least error-ed match from the above selections. 
\end{enumerate}
The method ensures that the matching error and the distance from the predicted point, both are considered for estimating pose of the vehicle.  
\section{Accelerometer Bias And Scale Factor Estimation} \label{Bias Estimation}

\subsection{Bias And Scale Factor Correction}
Inspired by the linear coordinate system introduced in Section \ref{Data Representation}, linear acceleration can be obtained by taking a second-order derivative of the distance traveled by the vehicle. 
Let $r_j$ be the position of the vehicle and $S_j$ be the total distance covered by the vehicle at the $j^{th}$ timestamp. 
\begin{align}
    {r_j} &= \begin{bmatrix}
        (x_j-x_{j-1}), & (y_j-y_{j-1})
    \end{bmatrix}^T + r_{j-1}\nonumber \\ 
    S_j &= \sqrt{(x_j-x_{j-1})^2 + (y_j-y_{j-1})^2} + S_{j-1} 
\end{align}
Utilizing the relationship between vehicle pose and traveled distance, $S = \sqrt{{r}^T r}$, the first derivative yields a relationship between linear resultant velocity along segment $S$ and the velocity vector $v_j$ as follows: 
\begin{equation}
    \frac{dS_j}{dt} = \frac{1}{S_j} {r_j}^T v_j
\end{equation}
Note that $\frac{dS_j}{dt}$ is $\frac{dS}{dt}$ evaluated at the $j^{th}$ timestamp. 
The second derivative yields a relationship between linear resultant acceleration along segment $S$ and acceleration vector $a_j$ as follows: 
\begin{align}\label{eq:sec_derS}
     \frac{d^2S_j}{dt^2} &= \frac{1}{S_j} ({v_j}^T v_j + {r_j}^T a_j - {\frac{dS_j}{dt}}^2)
\end{align}
Utilizing the accelerometer estimation model given in \eqref{eq:estimationModel}, \eqref{eq:sec_derS} can be rearranged in order to estimate the bias and scale factor of the accelerometer: 
\begin{align}
     {r_j}^T (k_a \tilde{a_j} -\beta_a) &= S_j\frac{d^2S_j}{dt^2} - {v_j}^T v_j + {\frac{dS_j}{dt}}^2  
    \label{eq:Resultant}
\end{align}
Similarly, two more equations can be obtained to relate the accelerations along $X$ and $Y$ axes with the distances traveled along $X$ and $Y$ axes as follows:
\begin{align} \label{eq:XDir}
\centering
    x_j-x_{j-1} &= {r_j}_x-{r_{j-1}}_x & y_j-y_{j-1} &= {r_j}_y-{r_{j-1}}_y\\ \nonumber
    \hat{{a_j}_x} &= \frac{d^2{r_j}_x}{dt^2} &  \hat{{a_j}_y} &= \frac{d^2{r_j}_y}{dt^2} \\ \nonumber
    {k_a}_x \tilde{a}_x - {\beta_a}_x &= \frac{d^2{r_j}_x}{dt^2} & {k_a}_y \tilde{a}_y - {\beta_a}_y &= \frac{d^2{r_j}_y}{dt^2}
\end{align}
Equations (\ref{eq:Resultant}), and (\ref{eq:XDir}) are combined into a least-square estimation system to determine the bias and scale factor of the accelerometer. The least-square problem is set in \eqref{eq:LeastSq}. The system uses sequential readings of the poses obtained from the magnetic field matching for better accuracy. In the next sections, the estimated scaling factor and bias are denoted by $(\Bar{k}_a, \Bar{\beta}_a)$.  
\begin{equation}
\label{eq:LeastSq}
\begin{bmatrix} \begin{smallmatrix}
    r_{1_x} a_{1_x} & r_{1_y} a_{1_y} & -r_{1_x} & -r_{1_y} \\
     . & . & . & . \\
     . & . & . & . \\
     r_{n_x} a_{n_x} & r_{n_y} a_{n_y} & -r_{n_x} & -r_{n_y} \\
     a_{1_x} & 0 & -1 & 0 \\
     . & . & . & . \\
     . & . & . & . \\
     a_{n_x} & 0 & -1 & 0 \\
     0 & a_{1_y} & 0 & -1 \\
     . & . & . & . \\
     . & . & . & . \\
     0 & a_{n_y} & 0 & -1 \\
\end{smallmatrix} \end{bmatrix}
\begin{bmatrix} \begin{smallmatrix}
    \Bar{k}_{a_x} \\
    \Bar{k}_{a_y} \\
    \Bar{\beta}_{a_x} \\
    \Bar{\beta}_{a_y}
\end{smallmatrix} \end{bmatrix} = 
\begin{bmatrix} \begin{smallmatrix}
    S_1\frac{d^2S_1}{dt^2} - {v_1}^T v_1 + ({\frac{dS_1}{dt}})^2  \\
    . \\
    . \\
    S_n\frac{d^2S_n}{dt^2} - {v_n}^T v_n + ({\frac{dS_n}{dt}})^2 \\
    \frac{d^2r_{1_x}}{dt^2} \\
    . \\
    . \\
    \frac{d^2r_{n_x}}{dt^2} \\
    \frac{d^2r_{1_y}}{dt^2} \\
    . \\
    . \\
    \frac{d^2r_{n_y}}{dt^2}
\end{smallmatrix} \end{bmatrix}
\end{equation}
\subsection{Bias And Scale Factor Update}
To facilitate an EKF update step for incorporating the above estimated factors, the measurement equation, measurement residual and measurement matrix are determined as follows. Let the right hand side of (\ref{eq:Resultant}) be,
\begin{equation}
    m_j =  S_j\frac{d^2S_j}{dt^2} - {v_j}^T v_j + {\frac{dS_j}{dt}}^2 
\end{equation}
Utilizing the accelerometer estimation model in \eqref{eq:estimationModel}, the following two equations for measurement residual and measurement equation are obtained respectively. The  measurement residual is obtained using the estimated bias and scaling factor from the least square estimation \eqref{eq:LeastSq}, whereas the measurement equation is obtained using the predicted bias and scaling factor from the time update \eqref{timeUpdate}.
\begin{align}
\centering
    z &= \Bar{k}_a \Tilde{a} - \Bar{\beta}_a  &
    h(x) &= \hat{k}_a \Tilde{a} - \hat{\beta}_a
\end{align}
Assuming that the acceleration value is unchanged from the timestamp $j-1$ to the timestamp $j$, $\hat{a}_{j-1} = \hat{a}_{j}$; the following equations can be written in a matrix format as shown:
\begin{align}     
    r^T_{j-1} \hat{a}_{j} &= m_{j-1} \\
    r^T_{j} \hat{a}_{j} &= m_{j}
\end{align}

The measurement model is given by,
\begin{equation}
\centering
    h_{j}(x) = \hat{a}_{j} =
{\begin{bmatrix}
   r^T_{j-1} \\
   r^T_{j} 
\end{bmatrix}}^{-1}
\begin{bmatrix}
    m_{j-1} \\
    m_{j}
\end{bmatrix}
\label{MeasEq}
\end{equation}
where,
\begin{align}
    R &= \begin{bmatrix}
   r^T_{j-1} \\
   r^T_{j} 
\end{bmatrix} &
    M &= \begin{bmatrix}
    m_{j-1} \\
    m_{j}
\end{bmatrix}
\end{align}

The partial derivative of $M$ with respect to the position at time $j$ is calculated as follows:
\begin{align} 
    \frac{\partial m_j}{\partial p_j} &=  \frac{\partial S_j}{\partial p_j} \frac{d^2 S_j}{dt^2} + \frac{\partial (\frac{d^2 S_j}{dt^2})}{\partial p_j} S_j - 2 \frac{d S_j}{dt}  \frac{\partial (\frac{d S_j}{dt})}{\partial p_j} \nonumber \\
    \frac{\partial M}{\partial p_j} &=  \begin{bmatrix}
    [0_{1\times2}] \\
    (\frac{\partial m_j}{\partial p_j})^T   
    \end{bmatrix}
\end{align}

The partial derivative for the entire expression is obtained by taking the derivative of multiplication of matrices as illustrated in \cite{MatMul}: 
\begin{equation} \small
    \frac{\partial(h(x_j))}{\partial p_j} = \frac{\partial R^{-1}}{\partial p_j} (M \circledast I_{2}) + R^{-1}  \frac{\partial M}{\partial p_j}   
\end{equation}

The partial derivative of (\ref{MeasEq}) with respect to the velocity at time $j$ is calculated as follows:
\begin{gather}
    \frac{\partial m_j}{\partial v_j} =  S_j \frac{\partial (\frac{d^2 S_j}{dt^2})}{\partial p_j} - 2v_j + 2 \frac{d S_j}{dt}  \frac{\partial (\frac{d S_j}{dt})}{\partial v_j} \nonumber \\
    \frac{\partial(h(x_j))}{\partial v_j} = R^{-1} \begin{bmatrix}
   0_{1\times2} \\
    (\frac{\partial m_j}{\partial v_j})^T 
\end{bmatrix} 
\end{gather}

The partials are put together to form the measurement matrix as shown in algorithm [\ref{alg:EKF}]. Note that, in algorithm [\ref{alg:EKF}] $x$, $P$, $Q$ and $R$ represent the state, covariance matrix, process error and measurement error.

\begin{algorithm} 
\caption{Extended Kalman Filter}\label{alg:EKF}
\begin{algorithmic}
\State \textbf{Time Update:}
\begin{align} \nonumber
\hat{X}_{n|n-1} &= f(X_{n-1},\omega,a) \\ \nonumber
\hat{P}_{n|n-1} &= F_n P_{n-1}F^T_n + GQG^T \nonumber
\end{align}

\State \textbf{Magnetic Measurement Update:}
\begin{align} \nonumber
h_{{mag}_{n|n-1}} &= \hat{p}_{n|n-1} \\ \nonumber
z_{{mag}_{n}} &= p_{s_{min}} \\ \nonumber
H &= \begin{bmatrix} 
\textbf{0} & \textbf{I} & \textbf{0} & \textbf{0} & \textbf{0} & \textbf{0} & \textbf{0} 
\end{bmatrix} \\ \nonumber
K_n &= \hat{P}_{n|n-1} H_n(H_n \hat{P}_{n|n-1} H^T_n + R)^{-1} \\ \nonumber
X_{n} &= \hat{X}_{n|n-1} + K_n (z_{{mag}_{n}} - h_{{mag}_{n|n-1}}) \\ \nonumber
q_n &= \hat{q}_{n|n-1} + \frac{1}{2} \Omega (\delta \alpha) \hat{q}_{n|n-1} \\ \nonumber
P_n &= (I-KH)\hat{P}_{n|n-1} \nonumber
\end{align}

\State \textbf{Accelerometer Measurement Update:} 
\begin{align} \nonumber
h_{{acc}_{n|n-1}} &= \hat{k}_a \Tilde{a} - \hat{\beta}_a \\ \nonumber
z_{{acc}_{n}} &= \Bar{k}_a \Tilde{a} - \Bar{\beta}_a \\ \nonumber
H &= \begin{bmatrix} 
\textbf{0} & \frac{\partial(h(x_j))}{\partial p_j} & \frac{\partial(h(x_j))}{\partial v_j} & \textbf{0} & \textbf{0} & \textbf{0} & \textbf{0}  
\end{bmatrix} \\ \nonumber
K_n &= \hat{P}_{n|n-1} H_n(H_n \hat{P}_{n|n-1} H^T_n + R)^{-1} \\ \nonumber
X_{n} &= \hat{X}_{n|n-1} + K_n (z_{{acc}_{n}} - h_{{acc}_{n|n-1}}) \\ \nonumber
q_n &= \hat{q}_{n|n-1} + \frac{1}{2} \Omega (\delta \alpha) \hat{q}_{n|n-1} \\ \nonumber
P_n &= (I-KH)\hat{P}_{n|n-1} \nonumber
\end{align}

\end{algorithmic}
\end{algorithm}


\section{Results}
\subsection{Data Collection}
\begin{figure}
\centering
\includegraphics[scale=0.23]{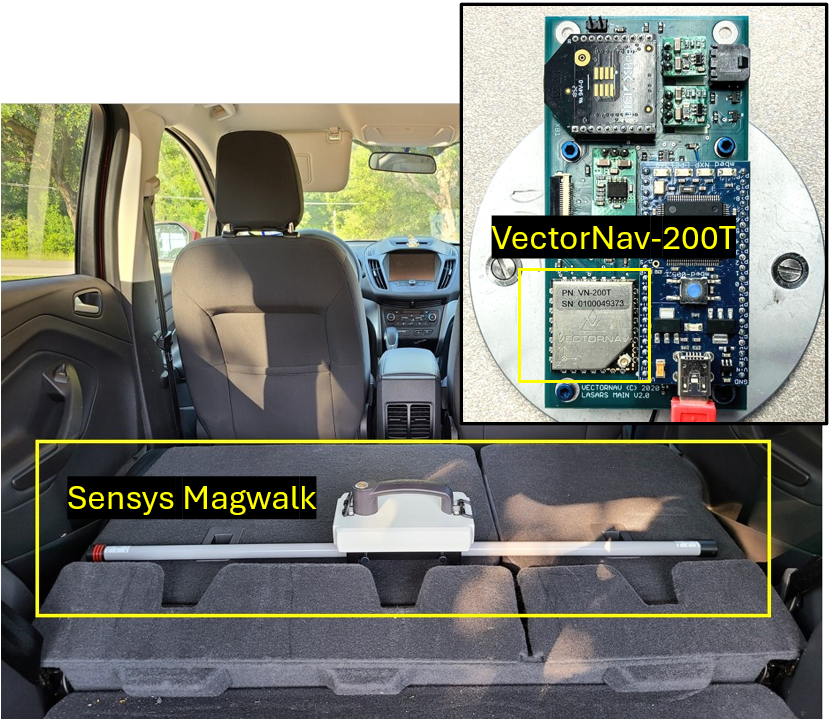}
\caption{Sensor Setup: The Magwalk magnetometer and VectorNav IMU are both places inside a vehicle as shown.}
\label{fig:setup}
\end{figure}
\begin{figure}
\centering
\includegraphics[scale=0.5]{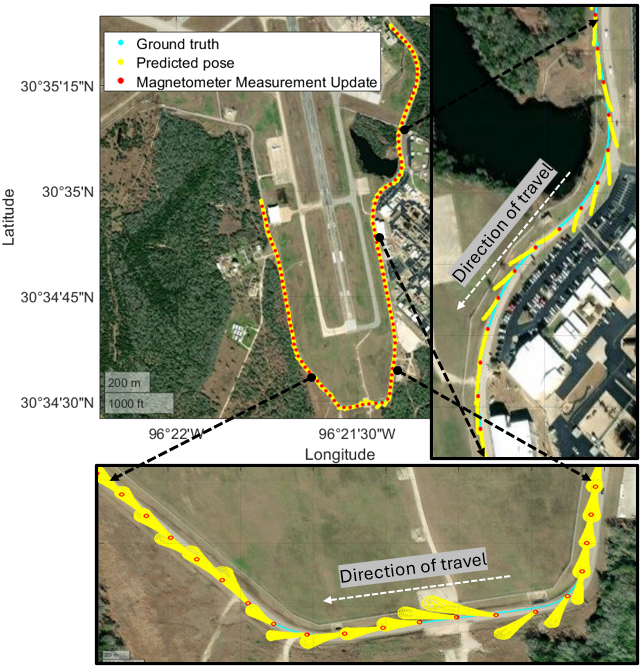}
\caption{Position estimation using EKF: The pose estimation is performed along a $3 \ Km \ long \ route \ in \ College \ Station, \ Texas$. The highlighted figures show the predicted and updated pose and covariance along the trajectory of the vehicle.}
\label{fig:Route}
\end{figure}

In order to demonstrate the capabilities of the proposed algorithm, it is tested on data collected by driving a car around suburban area. The data is collected using a lightweight, portable magnetometer survey kit from Sensys, namely, Magwalk\footnote{\url{https://sensysmagnetometer.com/products/magwalk/}} to measure the ambient magnetic field and a VectorNav-200T IMU unit\footnote{\url{https://www.vectornav.com/products/detail/vn-200}} to get the accelerometer and gyroscope measurements. Magwalk has two high-precision 3-axis magnetometers attached on the ends of a rod with magnetic resolution up to $0.3$ nanotesla ($n T$). We use one magnetometer at one end of the rod to collect the magnetic field data. The sensors are placed in a car as shown in Fig. \ref{fig:setup}. The data is collected at an interval of $100ms$ and the car is driven at around $28-32$ $Km/h$ on a $3 \ Km$ long route. The route taken for this test is shown in Fig. \ref{fig:Route}. For map building, the sensor data is collected with ground truth GPS locations.  

\subsection{Magnetic Field Map}
\begin{figure}
\centering
\includegraphics[scale=0.3]{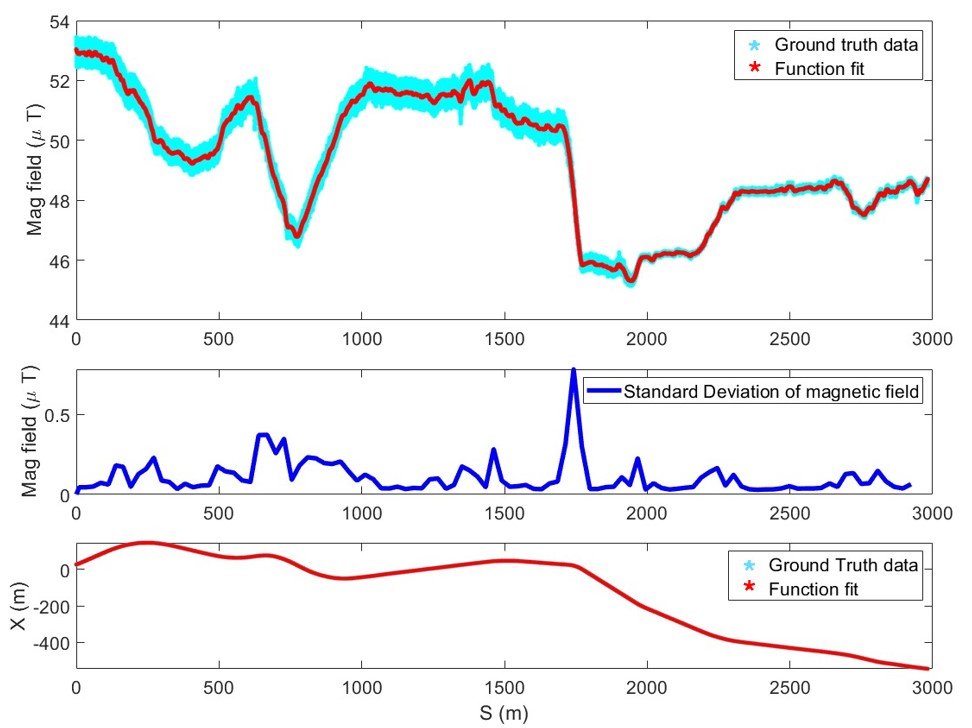}
\caption{Data Analysis: The first plot represents magnetic field data against the trajectory $S$ of the vehicle; the global function fit is obtained using segment-wise local fits. The second plot shows the standard deviation of the batches of magnetic field used for localization. The third plot illustrates the $X$ vs $S$ data and function fit. (As the pose data is sparse, it is not visible in the plot.) A similar plot for $Y$ vs $S$ is also a part of the map structure.   }
\label{fig:Map structures}
\end{figure}
The GPS locations of the vehicle are converted to a local coordinate system $(x,y)$ along with the traveled distance of the vehicle $S$. As described in Section \ref{Data Representation}, the local and global function fits for (the magnitude of the magnetic field vector vs $S$), (the $x$ position of the vehicle vs $S$), and (the $y$ position of the vehicle vs $S$) are obtained. Two of these curve fits are shown in the Fig. \ref{fig:Map structures} along with the magnetic field variations. The segments for local function fit are chosen to be of $10 \ m$, in order to capture the adequate amount of details of the curve.    

\begin{figure}
\centering
\includegraphics[scale=0.31]{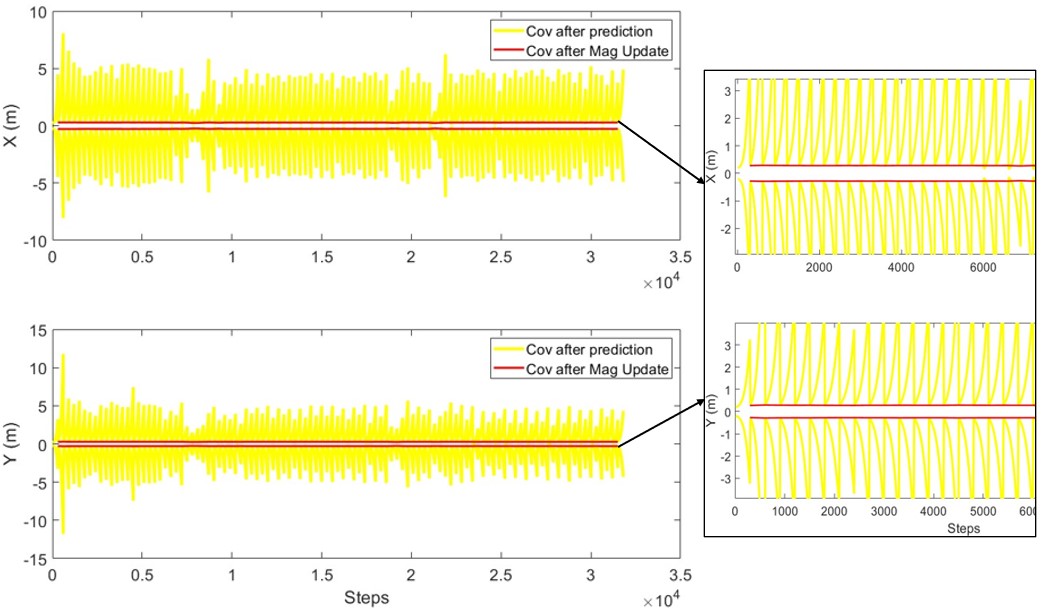}
\caption{Covariance analysis: The plot shows decrease in the pose covariance value after every magnetic field measurement update.}
\label{fig:Cov}
\end{figure}

\subsection{EKF testing}
The EKF testing is carried out such that the time update, magnetic measurement update, and accelerometer measurement update are performed at an interval of $0.1 \ sec, \ 3 \ secs,\  and \ 6 \ secs$ respectively. The update rates can be changed based on the diversity of the magnetic field around the area and the vehicle's speed. For the purpose of testing, the real-time magnetic field measurement values are simulated by introducing noise of the order of 0.1 microtesla $(\mu T)$ in the magnetic field fitted data points. Fig. \ref{fig:Route} shows estimated positions of the vehicle along the path for every prediction and magnetic measurement update step. The average deviation of the estimated poses from the ground truth poses is $0.47 \ m$. The covariance analysis in Fig. \ref{fig:Cov} demonstrates the reduction in pose uncertainty after every magnetic field measurement update. Note that the EKF is designed to rely mainly on magnetic field batch matching for position updates. 
\begin{figure}
\centering
\includegraphics[scale=0.24]{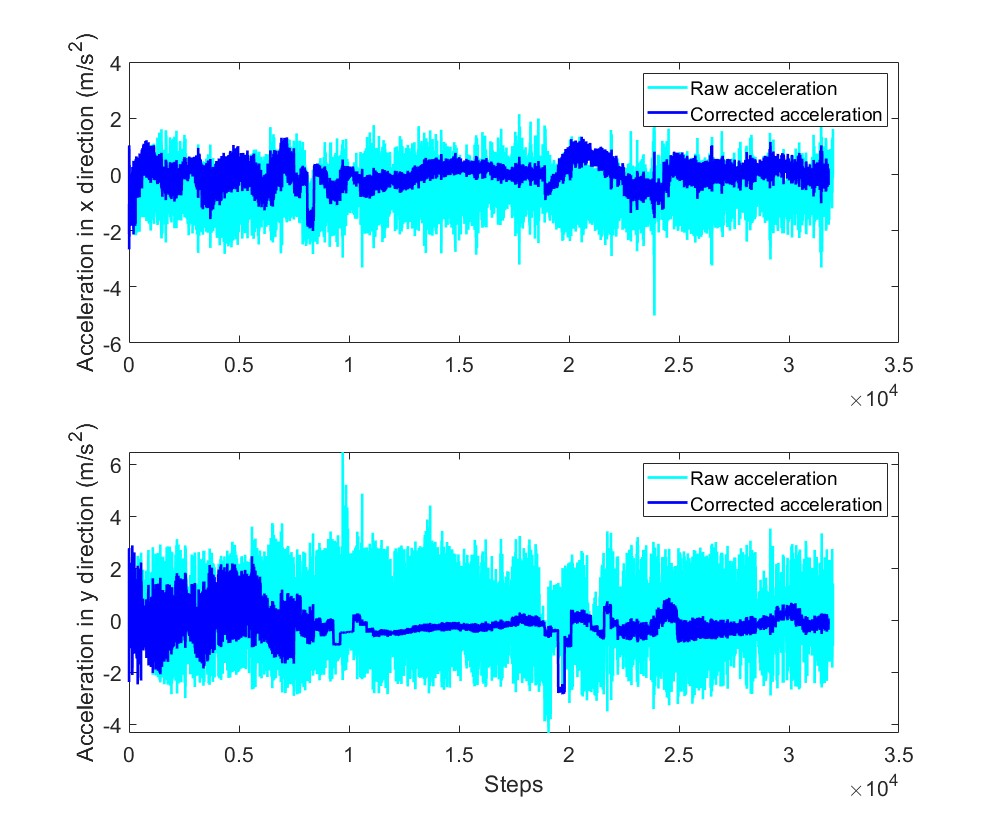}
\caption{Updated acceleration values plotted with the raw accelerometer values: The updated values show a less noisy, near-zero acceleration for a vehicle driving at a near-constant speed.}
\label{fig:AccelVals}
\end{figure}
The accelerometer bias and scale factors are estimated in the accelerometer measurement update step of the EKF. Fig. \ref{fig:AccelVals} shows raw acceleration values from the accelerometer and the updated acceleration values obtained from the estimated acceleration bias and scale factors. Note that the updated acceleration is less noisy as well as the values are trying to approach zero, representing the near-constant speed of the vehicle. 

\section{Conclusion}
This work establishes a foundation for a novel method to utilizing ambient magnetic field for mapping and localization. In-loop acceleration correction presents a promising method to reduce noise in accelerometer data. The use of mathematical function fits for the magnetic field data proves to be an efficient map structure, enabling the representation of large, continuous datasets with only a few coefficients. Moving forward, we plan to expand this method in two key areas. First, we intend to refine the data matching process to account for complex speed varying suburban environments, accounting for the changes in incoming data frequency. Second, we aim to extend the global functional mapping to 2D and 3D, facilitating localization in indoor, outdoor above-ground, or underwater settings. 

\section*{Acknowledgement}
We thank the U.S. National Science Foundation (NSF) for supporting the presented work under the grant 2137265 for IUCRC Phase I Texas A\&M University, Center for Autonomous Air Mobility and Sensing (CAAMS)\footnote{\url{https://www.nsf.gov/awardsearch/showAward?AWD_ID=2137265&HistoricalAwards=false}}.

\bibliographystyle{plain}
\bibliography{main.bib}

\addtolength{\textheight}{-12cm}   

\end{document}